\documentclass[11pt]{article}
\usepackage{graphicx}

\title{\textbf{Figuring out Actors in Text Streams: Using Collocations to establish Incremental Mind-maps}}
{\author{\small T. Rothenberger, S. \"Oz, E. Tahirovic\\
\small Johann Wolfgang Goethe-University Frankfurt am Main, Dept. of Computer Science\\
\small Robert-Mayer-Str. 11-15, 60486 Frankfurt am Main, Germany\\ 
\small Email: \{tahirov, oez, rothenb\}@cs.uni-frankfurt.de
\and
\small C. Schommer\\ 
\small University of Luxembourg, Dept. of Computer Science and Communication\\ 
\small 6, Rue Coudenhove-Kalergi, L-1359 Luxembourg\\  
\small Email: christoph.schommer@uni.lu
}

\date{\today}

\begin{document}
\maketitle

\begin{abstract}
The recognition, involvement, and description of main actors influences the story line of the whole text. This is of higher importance as the text per se represents a flow of words and expressions that once it is read it is lost. In this respect, the understanding of a text and moreover on how the actor exactly behaves is not only a major concern: as human beings try to store a given input on short-term memory while associating diverse aspects and actors with incidents, the following approach represents a virtual architecture, where collocations are concerned and taken as the associative completion of the actors' acting. Once that collocations are discovered, they become managed in separated memory blocks broken down by the actors. As for human beings, the memory blocks refer to associative mind-maps. We then present several priority functions to represent the actual temporal situation inside a mind-map to enable the user to reconstruct the recent events from the discovered temporal results.   
\end{abstract}

 \section{Motivation}
The main idea is to try to differentiate sentences on the basis of actors that appear in them.  We pursue \textit{one actor per sentence} rule to avoid ambiguity, so that each sentence becomes a part of the knowledge base for just one actor. Each actor who appears once receives his own mind-map. We reconstruct the story line of the text from these mind-maps to find out more about single actors by examining interaction among them. Because of an incremental nature, we examine one sentence per time in a subsequent manner while updating our mind-map continuously. In order to understand the motivation and the subject itself, it would be helpful to take some time to closer examine the corresponding concepts. Some of the concepts are subject of research in computational linguistics while others are more related to computer science. We interpret \textit{incremental} as we refer to its ability to gradually adjust to changes that happen over a period of time so that the present condition reflects these changes.

The central concept of this work is the concept of collocation, which is one of the most frequent stipulated concepts in natural language processing and computational linguistics. The difficulty to reconcile all aspects of collocations in one single definition has occupied many natural language experts and computer linguists over a large period of time. In \cite{KRE00}, The term \textit{collocation} \dots is used for word combinations that are lexically determined and constitute particular syntactic dependencies such as verb-object, verb-subject, adjective-noun relations,etc. The mind-map concept we would like to present in this paper bears resemblance to canonical concept of the associative mind-map, in a sense that the derivation a of mind-map address, for a particular collocation, results from the collocation's structure/formation itself. Questions similar to our task have already been raised in the domain of computer linguistic particularly with regard to \textit{information} and \textit{discourse} structure.

Most the work that has been done on this field is to present the semantical, temporal and psychological attributes of a discourse in a way that is interpretable by a machine. This differs from our task in the sense that features of our discourse analysis will be presented to a human user who will be hopefully able to get a good picture of the discourse considered. In \cite{HEY04}, a topic-tree model to form a discourse structure acceptable for further machine interpretation is introduced. They denote the focus of a sentence as a topic that corresponds corresponds to our notion of \textit{actor} or \textit{character}, who has to be identified for each sentence. In \cite{SCH04}, an associative mind-map is established that bases on artificial cells that communicate to each other and merge in case they represent the same content. Connections through other artificial cells are done by the Hebbian Learning principle, taking into account a forget in case the connection is less stimulated.

 \section{Architecture}
Given a text stream, we will identify the main actors in order to understand how they interact and concentrate on current events that have taken place in the shorter past. The interpretation of the concept \textit{shorter past} will be defined by a function in dependence to the position of the collocation in the text and the number of times where this collocation appears in the present moment. A detailed elaboration of the priority functions mentioned follows. The interaction between main actors can often reveal a coherence about the story line of the text. This brings us to the incremental nature of our work that permits an insight into the story line at an user-defined point of time. Each sentence has to pass a pre-processing step, which is done sentence-wide. Here, the sentence is brought to a form that supports the identification of the corresponding actor/character.

\subsection{Initializing the Mind-Map}
After having successfully identified the actor, the information in the sentence becomes part of the mind-map that is set up for this particular actor; additionally, we keep in mind the information about a story line that is broken down by actors in the text. The user interactively may decide the direction of action in the text. He will be presented the most newsworthy entries of the mind-map according to the already mentioned priority function, for all or for a couple of chosen actors. In this way,  we can spare the user from reading the whole text in search for a particular spot or in his efforts to get an understanding of what the text is about. The system is validated on fairy tales as these texts are simple regarding the linearity of the story line and the structure of the sentences. In particular, the tests and validation of the results will be carried out on the text of \textit{Little Red Riding Hood by the Brothers Grimm}.

Before we manage collocations in the mind-map of a particular actor, each text stream item is pre-processed. This does not violate the request of a stream as the collocation representation in the mind-map follows immediately. For this, we perform several pre-processing actions like stemming and parsing. This is not of too much freedom in rearranging the input stream. A more efficient identification of actors can be achieved if the text stream can be modified at once, therefore phenomenons like main clauses and subordinate clause, direct/reported speech, active/passive etc. are processed. We stem each word and identify each compound sentence while breaking them down into a more simple representation. With loss of information, we dissolve multiple sentences by deleting conjunctions and rely on some corrective actions by the human supervisor in the sense of human-computer interaction. However, for a directed or reported speech, it would be advisable to embed the statements that appear between quotations into the story. This could be accomplished by erasing quotation marks and omitting the notation about the speaker. Interrogative sentences as such carry no information relevant for our task, because it is quite difficult and therefore hard to determine: what is their contribution in forming the information structure of the text? For this, we believe to be wise to leave them out as a part of the pre-processing step.

To discover an unique form of the sentences and to transfer indirect sentences, we use again a stemmer, a particular grammar and the human supervisor who takes some corrective actions and who detects the correct reference between pronouns and the corresponding actors. Additionally, we manage the position of the sentence/collocation in the text by a simple counter that is incremented sentence by sentence. The following is an example, how the preprocessing step looks like. Following an orginal text like

{
\begin{itemize}
   \item[$\rightarrow$] Der Wolf legte sich wieder ins Bett und fing an zu schnarchen. Der J\"ager ging vorbei. Er dachte, die alte Frau schnarcht so laut, da muss ich einmal nachsehen. Er trat ein. Im Bett lag der Wolf, den er so lange gesucht hatte.
\end{itemize}
}

we dissolve the compound sentences incrementally. Word items are brought to their basic form, categories are assigned, and the correct references between pronouns and actors established.

{
   \begin{itemize}
        \item[$t_1$]: Der Wolf legte sich wieder ins Bett. $\Rightarrow$ Wolf(N) - legen(V) - Bett(N) 
        \item[$t_2$]: Der Wolf fing an zu schnarchen. $\Rightarrow$ Wolf(N) - anfangen(V) - schnarchen(N) 
        \item[$t_3$]: Der J\"ager ging vorbei. $\Rightarrow$ J\"ager(N) - gehen(V) - vorbei(N)
        \item[$t_4$]: Er dachte. $\Rightarrow$ Frau(N) - sein(V) - alt(ADJ)
        \item[$t_5$]: Die alte Frau schnarcht so laut. $\Rightarrow$ Frau(N) - schnarchen(V) - laut(ADJ)
        \item[$t_6$]: Ich muss einmal nachsehen. $\Rightarrow$ J\"ager(N) - nachsehen(V) - Haus(N)
        \item[$t_7$]: Er trat ein. $\Rightarrow$ J\"ager(N) - eintreten(V) - Haus(N)
        \item[$t_8$]: Im Bett lag der Wolf. $\Rightarrow$ Bett(N) - liegen(V) - Wolf(N)
        \item[$t_9$]: Er hatte den Wolf so lange gesucht. $\Rightarrow$ J\"ager(N) - suchen(V) - Wolf(N)
   \end{itemize}
}

\subsection{Assign collocations to an actor}

To assign each sentence to one particular actor, the canonical sentence/collocation structure is made up of several parts:

\begin{itemize}
     \item A sentence starts with a subject representing the actor (to whom the sentence is assigned).
     \item It is followed by a verb in the sentence that indicates the relation between the actor and the third part of the collocation.
     \item The third part is a generalized notion of an object in the sentence, which is involved in some way with the actor. It is normally a noun or an adjective depending on the way, how the sentence/collocation in question is generated in the pre-processing.
     \item At the end of each collocation, a position counter supports the process of representing collocations in a correct order.
\end{itemize}

The collocation that ends with an adjective is created from a word combination \textit{adjective + noun} in the original text. The \textit{adjective} describes one particular feature of the \textit{noun}; this yields on a new collocation in which \textit{noun} becomes the \textit{actor} and \textit{adjective} an \textit{object} of the collocation. We assign the word \textit{sein} a linking role in the collocation and represent it as a fact. The new collocation inherits the collocation number of the sentence, which it was a part of  and thus preserves the time flow of the text. This way we achieve the unique sentence structure that we need, and keep the original semantical value of the text.

\begin{center}\textit{alte Frau} $\Rightarrow$ \textit{Frau} - \textit{sein} - \textit{alt}\\
                      \textit{Blumen standen ringsherum} $\Rightarrow$ \textit{Blume} - \textit{stehen} - \textit{ringsherum}\end{center}

We keep a \textit{dynamic list} of each \textit{actor} who appears in text. Each \textit{actor} has one main mind-map block and one priority list for each priority function. Priority lists save collocations sorted by respective priority function enabling output in accordance with the particular function. These functions depend, in general, on the present moment and the repetition behavior of a particular collocation. In this respect, our strategy is as follows:

\begin{itemize}
    \item We identify the \textit{actor} in the currently processed collocation.
    \item We prove if this is the first appearance in the text. This can be achieved by contacting the dynamic collective \textit{actor} list. If so, a new mind-map block is allocated for this \textit{actor}. If the actor already exists in the mind-map, we then use the \textit{verb} (second part) of the collocation as a key to save this collocation in his main mind-map block. It is possible that this \textit{verb} appears in relation with this \textit{actor} again: then a list for it already exists in the mind-map. Such a list is kept for each \textit{verb}: what we still have to do is to examine if the \textit{object} (third part) can be found in the list. If so, we detect a reoccurrence of the current collocation.
    \item Further steps are to re-compute the priority functions and to sort the priority lists according to the new values. We preserve this way the correct order in the lists  which we later use in output for chosen actors.
\end{itemize}

\subsection{Priority Functions}
The priority lists is managed for each \textit{actor}. The presentation of the results - showing recent/current collocations broken down by actors - occurs directly from these priority lists. Each priority list has its own priority function by which the collocations become sorted. Dependent on the user's decision (who may intervene time by time), items of chosen actors are exported from the priority lists. However, not all collocations can be presented as a result. Since we try to model the current events in the story line, we prefer of just taking the most \textit{recent} collocations.

To provide the user with insight into each actor throughout the entire story line, we present at least five entries. Additionally, we use a threshold $\Delta$ to allow the output of the most recent collocations according to the priority function and display those entries in the actor's priority list that lie above. The results for a particular actor are presented  in a separate window, whereas the font size in which the results are written depends on the value evaluated by the priority function.

With the idea of priority functions, we model a concept of oblivion in the program. Taking the values computed by priority functions, we obtain an order among the collocations. For further elaboration it is important to define the notion of \textit{recent} or \textit{current} events. Observing repetition of a particular collocation could increase the importance of the collocation for the story line. Besides chronological features of the collocations, we considered  the number of repetitions as a meaningful feature in determining the priority of the collocation. This means that the collocations which have several occurrences in the text, should receive a larger priority than those which occur only once. Of course there are many other features which could seem reasonable in calculating the priority of the collocation, like appearance of a main character in it. We incorporated the impact of repetitions in the calculation of two of our priority functions. All priority functions should share some important properties in order to model the story line primarily in accordance with its chronological aspects. Such a function must be defined on a subset $D$ of the set of natural numbers $N$. It must be continuous and monotonously decreasing:

\begin{center}
$\forall x, y \in D : x < y \Rightarrow f(x) > f(y)$
\end{center}

We provide several priority functions from which the user can choose: let $\vec{{x}}^{k} \in N^d $ be a dynamic vector of all sentence numbers in which the collocation $k$ appears, d is at most the number of all sentences in the text
\begin{center}$F_1(c,\vec{x}^{k}) = \sum_i{0.5^{c-{{x^k}_i}}}$\end{center}
where $c$ the number of the sentence which is currently examined. We can see that this function incorporates a repetition characteristic of a collocation in the calculation of the priority. The most recent occurrences of the collocation carry more weight in the sum than those which lie far in the past. Because we chose the geometric progression the earlier occurrences will be \textit{forgotten} very fast. If a occurrence happened in a very near past it has a relatively big influence on the whole sum. The coefficient of the geometric progression remains 0.5 all through the process. 

The second priority function considers the number of repetitions only ($d$ the dimension of the vector $\vec{{x}}^{k}$ of the first function). In this case we increase with each repetition the coefficient of the geometric progression. The priority is calculated in dependence of the current coefficient $\widehat{a}$, number of the currently examined sentence $c$ and the number $l_k$ of most recent sentence  in which a collocation $k$ appeared:
\begin{center}
$\widehat{a} = \sum_{i=1}^{d}{{0.5^i}}$
\end{center}

Since we increase the coefficient, with each repetition, priority of the collocations that experience repetitions is uprated. We can see that the step of increase for the coefficient is calculated as a sum of a geometric progression with a start coefficient 0.5. This function rates repetitions higher than the first one since with each repeated occurrence the coefficient changes permanently  and the chronological order of repetitions plays no part any more. The second function is then:
\begin{center}
$F_2(\widehat{a},c,l_k) = \widehat{a}^{c-l_k}$
\end{center}

The third function is the simplest of all three. The repetitions are considered irrelevant. Priority depends on the current sentence number $c$ and the number $l_k$ of the last sentence with a particular collocation $k$ observed.
\begin{center}
$F_3(c,l_k) = 0.5^{c-l_k}$
\end{center}

\subsection{Example}
These functions provide a way to measure relevance of repetition for calculating the priority of collocations. Let $c=20$ and
\begin{itemize}
    \item[$\rightarrow$] \textit{Wolf - sein - b\"ose} (fifth sentence)
    \item[$\rightarrow$]\textit{Wolf - sein - b\"ose} (fifteenth sentence)
    \item[$\rightarrow$]\textit{Wolf - sein - b\"ose} (seventeen sentence)
\end{itemize}

then 

\begin{itemize}
  \item[$\rightarrow$] $F_1 = 0.5^{20-5}+ 0.5^{20-15}+ 0.5^{20-17} \approx 0.15628$
  \item[$\rightarrow$] $F_2 = (0.5^{1}+ 0.5^{2}+ 0.5^{3})^{20-17} \approx 0.66992$
  \item[$\rightarrow$] $F_3 = 0.5^{20-17} = 0.125$
\end{itemize} 

We can already see on this example how repetitions exert a different impact on the respective function evaluation.

\section{Implementation}
\begin{figure}[htbp]
   \centering
   \includegraphics[width=11cm]{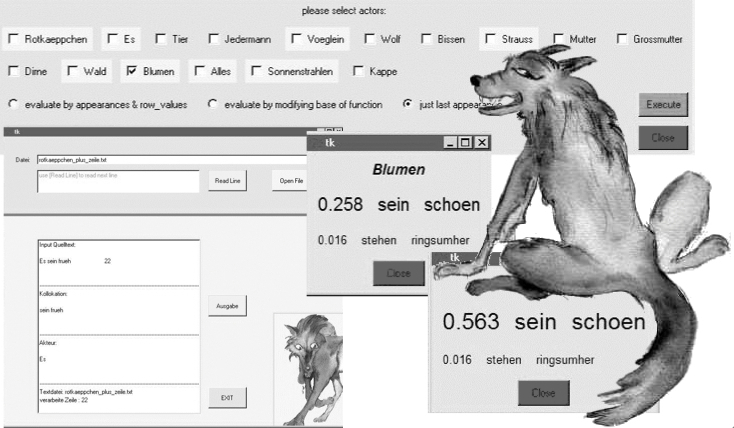}
   \caption{The situation for the actor \textit{Blumen} after having read the text stream. Depending on the priority function, the values differ, but the highest association keeps stable (\cite{ROT06}).}
   \label{fig:wolf}
\end{figure}

Figure \ref{fig:wolf} shows the workbench of the software system, which has been already presented in \cite{ROT06}. It consists of several panels: in the top, the user may select the discovered actors, which may be human beings, references to human beings like \textit{Jedermann}, or even animals. We then can select among the different priority functions receiving the mind-map with the associative relationships, including a relevance value. In this respect, Figure \ref{fig:wolf} shows the situation for the actor \textit{Blumen} after having read the text stream. The highest association is with \textit{sein sch\"on}. The priority values depend on the chosen priority function.

\section{Test and Validation}
\begin{figure}[htbp]
   \centering
   \includegraphics[width=13cm]{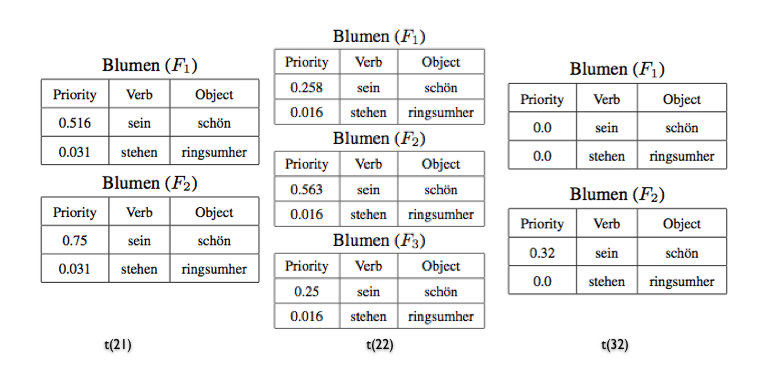} 
   \caption{\small The actor \textit{Blumen} and its associated collocations after having read the twenty-first sentence (left). The collocation \textit{Blumen - sch\"on} proves the highest value with priority function $F_2 = 0.563$ whereas priority functions $F_1 = 0.258$ and $F_3 = 0.25$ are lower. The collocations for \textit{Blumen} after having read the thirty-second sentence (right). The discrepancy between \textit{sch\"on} and \textit{ringsumher} is strongest for priority function $F_2$.}
   \label{fig:incr13}
\end{figure}

As an example, we examine the german fairy tale \textit{Rotk\"apchen} by taking the priority functions for each actor. With $c$, we express the current sentence number. In Figure \ref{fig:incr13}, we compare first the impact of unequal influence of repetition for two priority functions of the same actor. The collocation \textit{Blumen - sein - sch\"on} appears both in fifteenth as well as in twentieth sentence with 

\begin{center}
$F_1(21,(15,20)) < F_2(0.75, 21, 20)$
\end{center}

\begin{figure}[htbp]
   \centering
   \includegraphics[width=13cm]{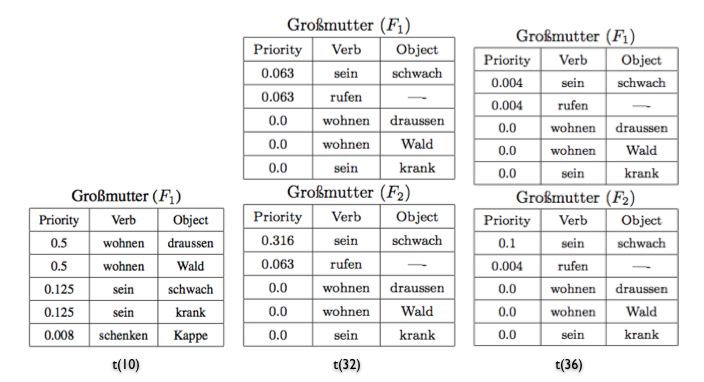} 
   \caption{\small The history of \textit{Gro\ss{}mutter} within the text at time points 10, 35, and 36. The word item \textit{wohnen} has disappeared (value of 0.0) as it has been forgotten whereas the association with \textit{schwach} weakly exists. The association \textit{Gro\ss{}mutter - rufen} has no object as \textit{rufen} does not need one.}
   \label{fig:incr14}
\end{figure}

We may observe that the second function is stronger influenced by repetitions. The priority values for \textit{Blumen - stehen - ringsumher} are the same since this collocation occurred only once so far. In Figure \ref{fig:incr13}, all three functions for the same actor here. The third function obviously calculates priority without regard to repetition since

\begin{center}
$F_3(22,20) <F_1(21,(15,20)) < F_2(0.75, 21, 20)$
\end{center}

After having read the thirty-second sentence, the list of collocations has changed (see Figure \ref{fig:incr13}). Following the priority function $F_1$, all associated weights are 0.0  whereas priority function $F_2$ gives a value of 0.32 for \textit{Blumen - sein - sch\"on}.

A similar situation occurs for \textit{Gro\ss{}mutter} (see Figure \ref{fig:incr14}). $F_1$ \textit{forgets} collocations which occured more than once faster than $F_2$. Moreover, we can recognize the geometric progression of the priorities calculated by $F_1$ for the actor \textit{Rotk\"appchen}. None of these collocations occurred more than once. The last of them (or the first by priority) occurred in the sentence number 32 with

\begin{center}
$F_1(32,(32))= 0.5^{32-32} = 0.5^0 = 1$
\end{center}

Once having read the text stream, the main memory for \textit{Wolf} is

{\small
\begin{itemize}
\item gehen: [[Wald, Gro\ss{}mutter] [Wald, 0.6, [10, 15]] [Gro\ss{}mutter, 0.5, [18]]] with a priority list of
   \begin{itemize}
          \item[$\rightarrow$] sein - b\"ose - 0.5
          \item[$\rightarrow$] sein - hungrig - 0.4
          \item[$\rightarrow$] sein - listig - 0.3
          \item[$\rightarrow$] suchen - Nahrung - 0.2
          \item[$\rightarrow$] ansprechen - Rotk\"appchen - 0.1
    \end{itemize}
\end{itemize}
}

\section{Conclusions}
In this paper, we have presented an architecture of revealing the action/story line in texts by recovering collocations between the actors.This has been done incrementally as the text can be seen as a text stream that once it is read it is lost. Bringing all sentences on a unique form allows the identification of the actor in a particular sentence. The number of actors per sentence is limited to one. The collocations corresponds to each involved action of an actor; this is managed by a mind-map allocated for each actor. A set of priority functions sort the collocations in each actor's mind-map by their actuality. An user may then decide which part and above all when he wants to check the story line. The output result presents the most actual collocations that actors are involved in; from these, the user may conclude how the actors interact with each other.

\section{Acknowledgement}
This work has been performed at the research laboratory \textit{ILIAS - Intelligent and Adaptive Systems} within the project \textit{TRIAS}, which is funded by the University of Luxembourg.


{\small
\begin{thebibliography}{4}
\bibitem{HEY04} G. Heyer, M. Läuter, U. Quasthoff, T. Wittig, C. Wolff: Learning Relations using Collocations. In: Maedche, Alexander; Staab, Steffen; Nedellec, C.; Hovy, Ed (ed.). Proceedings of IJCAI Workshop on Ontology Learning, Seattle/WA, August 2001, pp. 19-24.
\bibitem{KRE00}B. Krenn : CDB - A Database of Lexical Collocations. Austrian Research for Artificial Inteligence. 
\bibitem{ROT06}T. Rothenberger, S. \"Oz, E. Tahirovic: INASCO - Bestimmung eines inkrementellen Assoziativspeichers f\"ur Kollokationen. Internal Report. Johann Wolfgang Goethe Universit\"at Frankfurt am Main, 2006.
\bibitem{SCH04} C. Schommer: Incremental Discovery of Association Rules with Dynamic Neural nodes. Proceedings of the Workshop on Symbolic Networks. ECAI 2004, Valencia, Spain. 2004.
\bibitem{SCH07} B. Schroeder, M. Hilker, R. Weires: Dynamic Association Networks in Information Management. Proceedings 21st International Conference on Computer, Electrical, and Systems Science, and Engineering (CESSE 2007). Vienna, Austria, 2007.
\end{thebibliography}
}
\end{document}